\documentclass[conference]{IEEEtran}
\IEEEoverridecommandlockouts
\usepackage{cite}
\usepackage{amsmath,amssymb,amsfonts}
\usepackage{algorithmic}
\usepackage{graphicx}
\usepackage{textcomp}
\usepackage{xcolor}
\usepackage[utf8]{inputenc}
\usepackage[greek,english]{babel}
\usepackage{alphabeta}
\usepackage{soul}
\usepackage{subcaption}
\usepackage[export]{adjustbox}
\usepackage{booktabs}
\usepackage{makecell}
\definecolor{cream}{RGB}{255,253,240}
\renewcommand{\baselinestretch}{0.979}

\newcommand\freefootnote[1]{%
  \let\thefootnote\relax%
  \footnotetext{#1}%
  \let\thefootnote\svthefootnote%
}

\def\BibTeX{{\rm B\kern-.05em{\sc i\kern-.025em b}\kern-.08em
    T\kern-.1667em\lower.7ex\hbox{E}\kern-.125emX}}
\begin{document}

\title{GPlaceRL: An Open-Source Graph Reinforcement Learning Framework for Detailed Placement}

\author{\IEEEauthorblockN{Pavlos Stoikos\textsuperscript{†},~Foteini Oikonomou\textsuperscript{†},~Christos Poulos\textsuperscript{†},~Maria Pantazi-Kypraiou\textsuperscript{†},\\ ~Athanasios~Tziouvaras\textsuperscript{†}, Christos~Anagnostopoulos\textsuperscript{*},~Georgios~Karakonstantis\textsuperscript{†}~and~George~Floros\textsuperscript{‡}}

\IEEEauthorblockA{\textsuperscript{†}Department of Electrical and Computer Engineering, University of Thessaly, Volos, Greece\\
\{pastoikos, fotoikon, cpoulos, mpantazi-, attziouv, gkarakon\}@e-ce.uth.gr\\
\textsuperscript{*}School of Computing Science, University of Glasgow, UK, christos.anagnostopoulos@glasgow.ac.uk\\
\textsuperscript{‡}Department of Electronic and Electrical Engineering, Trinity College Dublin, Ireland, florosg@tcd.ie}}

\maketitle


\begin{abstract}
Reinforcement learning (RL) has emerged as a promising approach for placement optimization, particularly when combined with graph neural networks (GNNs) that capture circuit connectivity. However, most learning-based placement approaches focus on floorplanning, macro placement, or global placement, while detailed placement refinement remains relatively unexplored. In this paper, we present GPlaceRL, an open-source graph reinforcement learning framework for detailed placement refinement. GPlaceRL represents legalized placements as graphs and provides a modular environment for studying graph encoders, policy architectures, reward formulations, and local placement actions. To demonstrate the capabilities of GPlaceRL, we conduct a systematic evaluation of proximal policy optimization (PPO) policies with graph attention network (GAT) encoders in a per-design optimization setting.
Across five placement benchmarks, the best greedy evaluation results achieve HPWL improvements ranging from $3.27\%$ to $32.87\%$.
The results highlight the importance of compact GAT architectures and flexible local action spaces for placement optimization.
Overall, GPlaceRL provides a reproducible and extensible framework for systematic research on RL-based detailed placement refinement.

\end{abstract}

\begin{IEEEkeywords}
Open-Source Frameworks, Reinforcement Learning, Graph Neural Networks, Detailed Placement
\end{IEEEkeywords}

\section{Introduction}
Placement is a central stage of the physical design flow, directly affecting wirelength, timing, routability, and power. After global placement and legalization, detailed placement locally refines a legal placement through operations such as cell shifting, swapping, and reordering. Although these local changes may appear small, they can significantly affect half-perimeter wirelength (HPWL) and downstream routing quality~\cite{7999056}. 

\freefootnote{*This work was funded under the COIN-3D project, which has received funding from the European Union's Horizon Europe research and innovation program under grant agreement No. 101159667.}

Traditional detailed placement algorithms rely on carefully engineered local-search and optimization techniques, including dynamic programming, network flow, and min-cost matching formulations~\cite{1560039}. These methods are effective, but they often require evaluating many candidate local moves and may become costly as modern netlists grow in size and connectivity complexity~\cite{10.1145/1118299.1118353}.

Recent learning-based placement methods have shown that graph neural networks (GNNs) and reinforcement learning (RL) can capture structural information from netlists and optimize placement-related objectives~\cite{Mirhoseini2021AGP,9218582,9628153,10137100}. However, most RL-based works focus on floorplanning, macro placement, or global placement. In contrast, detailed placement refinement remains less explored, and open, reproducible frameworks for systematically studying RL-based detailed placement are still limited. This setting differs from other placement problems because the input is already legal, actions must preserve legality, and improvements arise from local refinements.

This work presents GPlaceRL~\cite{GPlaceRL}, an open-source graph reinforcement learning framework for detailed placement refinement. To demonstrate the capabilities of GPlaceRL, we study policies trained with proximal policy optimization (PPO) and graph attention network (GAT) encoders through a design-space exploration of architecture and action-space choices. This setup enables systematic evaluation of RL design choices for HPWL reduction. Rather than proposing a universal placer, we provide a reproducible framework for detailed placement research. The main contributions of this paper are summarized hereafter. \textit{First}, we introduce GPlaceRL, an open-source graph reinforcement learning framework for detailed placement refinement. \textit{Second}, we formulate detailed placement refinement as a per-design RL problem using graph-based placement states and legal local actions. \textit{Third}, we demonstrate the capabilities of GPlaceRL through a systematic design-space exploration of PPO policies with GAT encoders, investigating the impact of encoder architecture and action-space design. \textit{Finally}, we evaluate the framework across five placement benchmarks and analyze the resulting HPWL~improvements.

The remainder of the paper presents the GPlaceRL framework, evaluates the explored PPO-GAT configurations, and summarizes the main findings.

\section{GPlaceRL Framework}

\begin{figure*}[!htb]
    \centering
    \includegraphics[width=\textwidth]{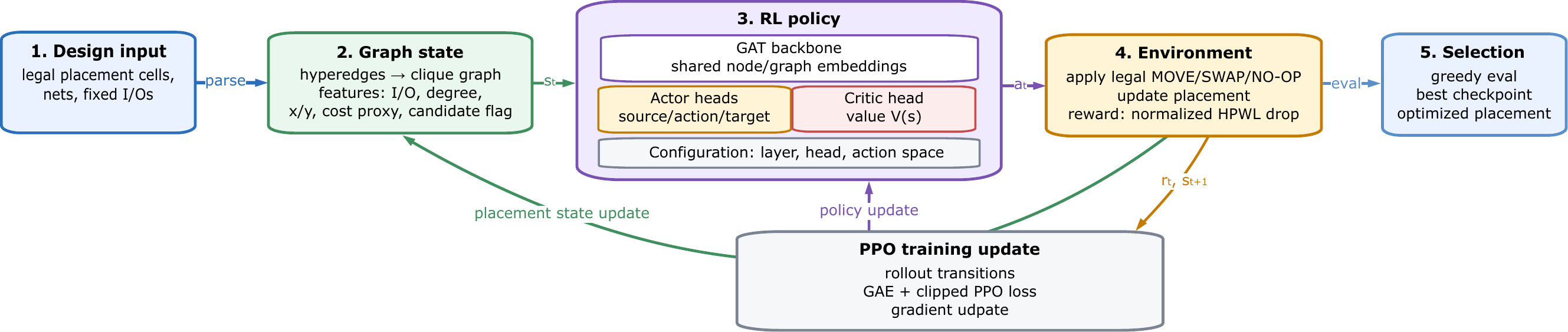}
    \caption{Overview of the GPlaceRL framework for detailed placement refinement. A legal placement is converted into a graph state and processed by a graph-based actor–critic policy. In the current implementation, a PPO-GAT policy applies local refinement actions, while periodic greedy evaluation selects the best checkpoint for generating the final optimized placement.}
    \label{fig:architecture}
\end{figure*}

GPlaceRL formulates detailed placement refinement as a sequential decision-making problem. Given an initial legal placement, the agent repeatedly applies local modifications and receives feedback according to the resulting change in HPWL. Rather than training a universal policy to generalize across unrelated circuits, GPlaceRL adopts a per-design optimization setting: for each benchmark, the policy is trained on the corresponding placement graph and used to produce a refined placement for that design. This allows the agent to exploit design-specific connectivity and spatial structure.

Figure~\ref{fig:architecture} illustrates the complete flow. The legal placement and netlist are parsed into a graph state that captures both connectivity and physical location. A PPO actor--critic policy with a shared GAT encoder processes this state and predicts a factorized action. The placement environment applies the selected local operation, updates the graph features, and returns an HPWL-based reward. During training, PPO updates the policy from sampled rollouts, while periodic greedy evaluation selects the checkpoint that produces the best placement.

\subsection{Graph representation of placement}

To expose circuit connectivity to the policy, the placement is represented as a graph $G=(V,E)$. Nodes correspond to movable standard cells and fixed I/O pins, while edges approximate net connectivity. Since physical nets are hyperedges, each net is converted into pairwise graph edges by connecting the components that belong to the same net. The resulting graph is processed using PyTorch Geometric. Each node $v$ is assigned the feature vector:
\vspace{-1pt}
\[
\mathbf{x}_v =
[\, \mathrm{is}_{io},\, \mathrm{deg}_{norm},\, x_{norm},\, y_{norm},\, \mathrm{cost}_{proxy},\, \mathrm{is}_{cand} \,]^{\top}
\]

The first two features are static. The binary feature $\mathrm{is}_{io}$ identifies fixed I/O pins, and $\mathrm{deg}_{norm}$ is the graph degree normalized by the maximum node degree. The remaining features are recomputed after each placement action. The normalized coordinates $(x_{norm},y_{norm})$ encode physical position inside the placement core. The feature $\mathrm{cost}_{proxy}$ is a local wirelength proxy, computed as the average Manhattan distance from a node to its graph neighbors and normalized by the core dimensions. Finally, $\mathrm{is}_{cand}$ marks whether a movable cell belongs to the current source-cell candidate set. Together, these features provide the policy with connectivity, geometry, and local placement-quality information.

\subsection{Placement environment and action space}

The environment operates on a placement grid with discrete legal sites. Each episode starts from the benchmark legal placement. The environment maintains legality throughout the trajectory by masking invalid choices and applying only actions that preserve placement constraints. The objective is to reduce HPWL, which is widely used as a proxy for routing cost during placement optimization.

Because the full placement decision space is extremely large, we use the factorized action representation shown in Fig.~\ref{fig:architecture}. At each step, the policy makes the following conditional decisions:

\begin{enumerate}
\item \textbf{Source cell selection:}  
A candidate set $\mathcal{C}$ is constructed from movable cells with the highest $\mathrm{cost}_{proxy}$ values. The agent first selects one source cell from this set.

\item \textbf{Action type selection:}  
Given the selected source cell, the agent selects one operation from the experiment-specific action set: \texttt{MOVE}, \texttt{SWAP}, or \texttt{NO-OP}. In the ablation study, this set is restricted to \textit{move-only}, \textit{swap-only}, or \textit{swap \& move} variants.

\item \textbf{Target selection:}  
Depending on the selected operation, the agent chooses either another cell for swapping or an empty site for relocation. Candidate targets are restricted to legal positions within a local spatial window around the source cell. If \texttt{NO-OP} is selected, the placement state is left unchanged.
\end{enumerate}

This formulation reduces the action-space size while retaining the two local transformations needed for detailed refinement: small relocations and pairwise exchanges.

\subsection{Policy network}

The policy follows an actor--critic architecture with a shared GAT encoder. The encoder processes node features and graph connectivity using stacked GAT layers, producing node embeddings and a global graph embedding obtained through mean pooling.

The actor is divided into three heads that mirror the factorized action. The source head scores candidate cells, the action head predicts the refinement operation, and the target head scores valid target cells or sites conditioned on the selected source and operation. The critic estimates the state value $V(s_t)$ from the global graph embedding.

This architecture is well suited for placement refinement because attention allows the encoder to weight neighboring cells and pins according to their relevance, rather than treating all graph connections uniformly.

\subsection{Training and reward}

The policy is trained using PPO, which stabilizes learning by constraining policy updates through a clipped objective. During rollouts, the agent samples actions from the factorized policy distributions and collects transitions $(s_t, a_t, r_t, s_{t+1})$. The critic estimates $V(s_t)$, and advantages are computed using generalized advantage estimation (GAE) to reduce variance in the policy-gradient estimates.

The reward is defined using the incremental improvement in HPWL:
\vspace{-1pt}
\begin{equation}
\label{eq:reward}
r_t = \frac{\mathrm{HPWL}_{t-1} - \mathrm{HPWL}_t}{\mathrm{HPWL}_{start}}
\end{equation}
\noindent
, where $\mathrm{HPWL}_{start}$ is the wirelength of the initial legal placement. A positive reward indicates that the current action reduces HPWL, while a negative reward penalizes degradations. Normalizing by the initial HPWL makes rewards comparable across designs with different wirelength scales. This dense reward provides immediate feedback after each local refinement step and encourages the agent to discover beneficial action sequences.

\subsection{Evaluation}

During evaluation, the learned policy is applied greedily. At each step, the highest-scoring legal action is selected instead of sampling from the policy distribution. Since a later action may partially undo an earlier improvement, we report the best HPWL reached during the evaluation trajectory:
\begin{equation}
\label{eq:eval_delta}
\Delta(\%) = 100 \cdot
\frac{\mathrm{HPWL}_{start} - \mathrm{HPWL}_{best}}
{\mathrm{HPWL}_{start}}
\end{equation}
The checkpoint achieving the best median $\Delta(\%)$ across evaluation rollouts is selected as the final optimized policy.


\section{Experimental evaluation}

\subsection{Experimental setup}

We evaluate GPlaceRL using PPO-GAT policies on five internally generated placement benchmarks representative of modern digital designs, named as B1--B5. Each benchmark corresponds to a legal placement snapshot produced by an EDA flow and contains the placement core, standard cells, fixed I/O pins, and net connectivity. Therefore, the problem considered in this work is not initial placement, but \emph{placement refinement}: the agent starts from an already legal placement and applies local actions to reduce wirelength while preserving legality.
Table~\ref{tab:benchmarks} summarizes the main characteristics of the evaluated benchmarks.

\begingroup
\renewcommand{\arraystretch}{1.0}
\begin{table}[!htb]
\caption{Statistics of the evaluated placement benchmarks.}
\label{tab:benchmarks}
\centering
\resizebox{\columnwidth}{!}{
\begin{tabular}{cccccc}
\toprule
Design & Core size & Cells & Nets & IOs & Initial HPWL \\
\midrule
B1 & $9.27\times9.216$ & 85 & 1494 & 26 & 1006 \\ 
B2 & $21.78\times21.888$ & 437 & 12128 & 55 & 13028 \\ 
B3 & $54.0\times54.144$ & 2976 & 135314 & 632 & 206641 \\ 
B4 & $76.14\times76.032$ & 6848 & 372952 & 332 & 723181 \\
B5 & $129.24\times129.024$ & 19463 & 1010888 & 674 & 3469054 \\
\bottomrule
\end{tabular}}
\end{table}
\endgroup

To study the effect of the model configuration, we conduct both an ablation study and a design-space exploration. More specifically, we vary the number of GAT layers ($\mathrm{L} \in \{2,4\}$), the number of attention heads ($\mathrm{H} \in \{4,8\}$), and the available action space of the PPO agent. Three action settings are examined: \textit{move-only}, \textit{swap-only}, and \textit{swap \& move}. All experiments are trained for 1500 episodes with batch size 10 and learning rate $10^{-4}$. As performance metric, we report the best greedy HPWL improvement $\Delta(\%)$ defined in Eq.~(\ref{eq:eval_delta}).

\subsection{Experimental results}

Table~\ref{tab:best_results} summarizes the best evaluation result obtained for each benchmark and action space, together with the corresponding GAT configuration. The reported value corresponds to the best greedy evaluation checkpoint observed during training, rather than necessarily the final training episode. Across the five benchmarks, the best HPWL improvement ranges from $3.27\%$ on B5 to $32.87\%$ on B1. The combined \textit{swap \& move} action space achieves the best result for every evaluated design, indicating that combining displacement and pairwise exchange operations gives the policy the most effective refinement capability. Although performance varies across benchmarks and architectures, the results reveal three consistent trends.

\begingroup
\renewcommand{\arraystretch}{1.0}
\begin{table}[!htb]
\caption{The top-performing RL architecture for each benchmark and action space.}
\label{tab:best_results}
\centering
\resizebox{\columnwidth}{!}{
\begin{tabular}{ccccc}
\toprule
Design & Action space & Best eval $\Delta$ (\%) & Best config & Best episode\\
\midrule
B1 & Move only & 28.26 & L2-H4 & 1400 \\ 
B1 & Swap only & 23.10 & L2-H4 & 60 \\ 
B1 & Swap + Move & \textbf{32.87} & L2-H4 & 1360 \\ \midrule
B2 & Move only & 12.65 & L2-H8 & 1380 \\ 
B2 & Swap only & 8.86 & L4-H8 & 70 \\ 
B2 & Swap + Move & \textbf{13.35} & L2-H8 & 1280 \\ \midrule
B3 & Move only & 10.58 & L4-H8 & 820 \\
B3 & Swap only & 7.94 & L4-H8 & 360 \\ 
B3 & Swap + Move & \textbf{12.11} & L2-H4 & 870 \\ \midrule
B4 & Move only & 3.96 & L2-H8 & 770 \\ 
B4 & Swap only & 1.74 & L2-H8 & 1260 \\ 
B4 & Swap + Move & \textbf{5.25} & L2-H4 & 420 \\ \midrule
B5 & Move only & 2.91 & L2-H8 & 1170 \\ 
B5 & Swap only & 0.18 & L4-H8 & 570 \\ 
B5 & Swap + Move & \textbf{3.27} & L2-H4 & 980 \\ 
\bottomrule
\end{tabular}
}
\end{table}
\endgroup

\textbf{GNN layers.} Table~\ref{tab:best_results} shows that the best overall
results are consistently obtained with two-layer GAT models. Four-layer
configurations appear only in a few action-specific cases, mainly when the
policy is restricted to simpler action spaces. This suggests that, for placement
refinement, shallow message passing provides sufficient local structural context
while avoiding the instability that deeper GNNs may introduce through
oversmoothing and over-squashing.

\textbf{Attention heads.} Increasing the number of attention heads does not necessarily translate into better results. Both H4 and H8 configurations appear among the best entries, depending on the benchmark and action space. For example, B4 selects an H8 configuration under \textit{swap-only}, whereas its best overall result is obtained with an H4 configuration under \textit{swap \& move}. This suggests that attention-head count should be treated as a design parameter rather than a monotonic source of improvement, which is consistent with prior observations on diminishing returns from additional heads~\cite{att_heads}.

\begin{figure*}[!htb]
  \centering
  \includegraphics[width=\textwidth]{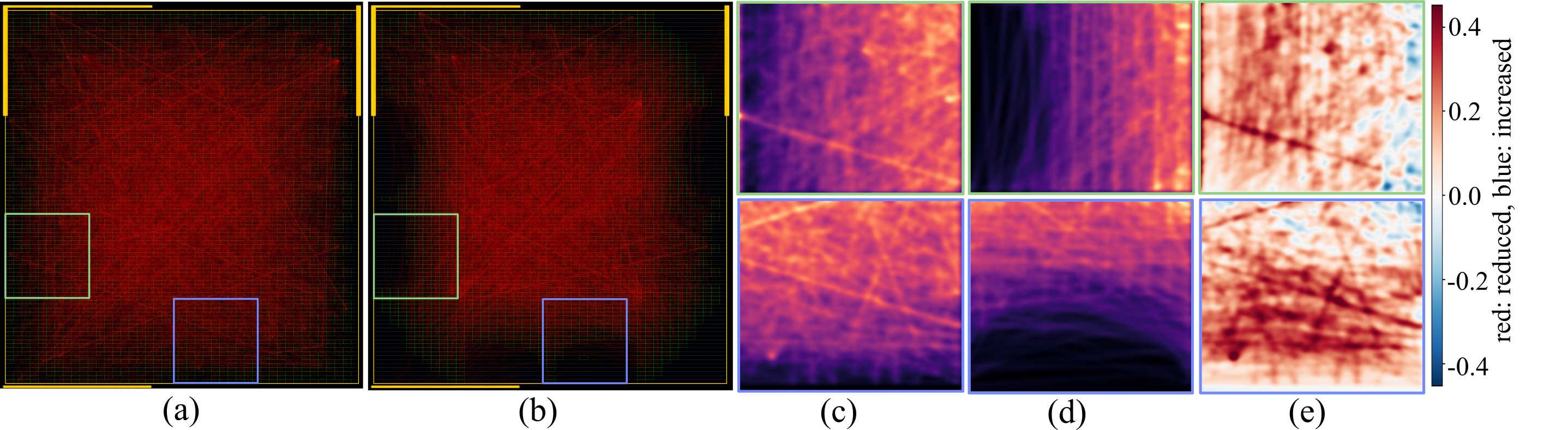}
  \caption{Qualitative placement and wire-density analysis for the best B3 result in Table~\ref{tab:best_results}, obtained with the \textit{swap \& move} action space and yielding a $12.11\%$ HPWL improvement. (a) Initial placement with selected local windows. (b) Refined placement after optimization. (c) Before-optimization wire-density maps for the selected windows. (d) After-optimization wire-density maps. (e) Density-change maps computed as before minus after, where red indicates reduced wire density and blue indicates increased wire density.}
  \label{fig:execute_window_density}
\end{figure*}

\textbf{RL action space.} The action space has the clearest impact on the final evaluation score. For all five benchmarks, \textit{swap \& move} outperforms both \textit{move-only} and \textit{swap-only}, with the largest absolute gains observed on B1 and B3. This behavior indicates that cell displacement and pairwise exchange are complementary: move actions provide fine-grained local adjustment, while swap actions can escape local ordering constraints. The benefit is especially visible on larger designs such as B3 and B4, although the corresponding convergence behavior remains noisier, as discussed below.

Figure~\ref{fig:execute_window_density} provides a qualitative view of the placement refinement achieved by the best B3 policy. This run corresponds to the \textit{swap \& move} entry in Table~\ref{tab:best_results} and achieves a $12.11\%$ HPWL improvement. The full-placement panels compare the design before and after optimization, while the highlighted windows focus on local regions where wire-demand changes are visually apparent. The corresponding density maps indicate that the refinement reduces wire concentration in selected regions, which is consistent with the measured HPWL reduction.

\begin{figure}[!t]
  \centering
  \includegraphics[width=\columnwidth]{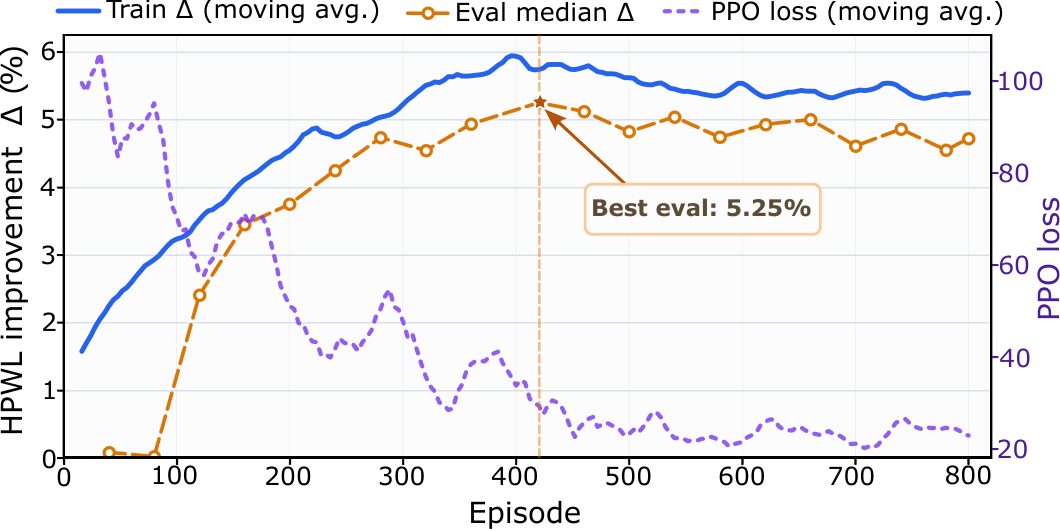}
  \caption{Convergence behavior for the B4 \textit{swap \& move} experiment. The best evaluation reaches a $5.25\%$ HPWL improvement at episode $420$, corresponding to the selected B4 result in Table~\ref{tab:best_results}.}
  \label{fig:div_convergence}
\end{figure}

Finally, Figure~\ref{fig:div_convergence} illustrates the convergence behavior observed for the B4 \textit{swap \& move} policy reported in Table~\ref{tab:best_results}. The best greedy evaluation reaches a $5.25\%$ HPWL improvement at episode $420$, matching the selected B4 result in the table. The training improvement increases rapidly during the early episodes and then reaches a plateau, while the greedy evaluation curve peaks and subsequently remains close to, but below, its best value. This trend indicates that the policy progressively identifies useful local refinements, with the PPO loss decreasing as training stabilizes.

\section{Conclusion}
\vspace{-2pt}
In this paper, we presented GPlaceRL, an open-source graph reinforcement learning framework for detailed placement refinement, and demonstrated its capabilities through a systematic evaluation of PPO-GAT policies. The proposed formulation operates directly on legal placements and applies local refinement actions while preserving legality. Across five benchmarks, the best greedy evaluation results achieve HPWL improvements ranging from 3.27\% to 32.87\%. The results show that compact two-layer GAT policies are generally the most effective and that combined swap-and-move action spaces outperform individual action variants. Overall, GPlaceRL provides a reproducible framework for systematic research on RL-based detailed placement refinement. Future work will focus on larger benchmarks and multi-objective optimization incorporating congestion and timing awareness.

\begingroup
\scriptsize
\bibliographystyle{IEEEtran}
\bibliography{bibliography}
\endgroup

\end{document}